\documentclass[11pt]{article}
\usepackage{eacl2014}
\usepackage{times}
\usepackage{latexsym}
\usepackage{amsmath}
\usepackage{multirow}
\usepackage{url}
\usepackage{graphicx}
\usepackage{subfig}

\DeclareMathOperator{\PYP}{PYP}

\newcommand{\hide}[1]{}

\title{Modelling the Lexicon in \\ Unsupervised Part of Speech Induction}
\author{Greg Dubbin \\
Department of Computer Science	\\
University of Oxford	\\
United Kingdom	\\
\tt{Gregory.Dubbin@wolfson.ox.ac.uk}\\\And
Phil Blunsom	\\
Department of Computer Science	\\
University of Oxford	\\
United Kingdom	\\
\tt{Phil.Blunsom@cs.ox.ac.uk}	\\
}
\date{}

\begin{document}
\maketitle

\begin{abstract}
Automatically inducing the syntactic part-of-speech categories for words in text is a fundamental task in Computational Linguistics. 
While the performance of unsupervised tagging models has been slowly improving, current state-of-the-art systems make the obviously incorrect assumption that all tokens of a given word type must share a single  part-of-speech tag. This one-tag-per-type heuristic counters the tendency of Hidden Markov Model based taggers to over generate tags for a given word type.  However, it is clearly incompatible with basic syntactic theory. 
In this paper we extend a state-of-the-art Pitman-Yor Hidden Markov Model tagger with an explicit model of the lexicon. 
In doing so we are able to incorporate a soft bias towards inducing few tags per type. 
We develop a particle filter for drawing samples from the posterior of our model and present empirical results that show that our model is competitive with and faster than the state-of-the-art without making any unrealistic restrictions.
\end{abstract}

\section{Introduction}
\label{sec:Introduction}

Research on the unsupervised induction of part-of-speech (PoS) tags has the potential to improve both our understanding of the plausibility of theories of first language acquisition, and Natural Language Processing applications such as Speech Recognition and Machine Translation.
While there has been much prior work on this task \cite{brown92class,clark03combining,christodoulopoulos10two,toutanova08pos,goldwater07pos,blunsom11acl}, a common thread in many of these works is that models based on a Hidden Markov Model (HMM) graphical structure suffer from a tendency to assign too many different tags to the tokens of a given word type.  
Models which restrict word types to only occur with a single tag show a significant increase in performance, even though this restriction is clearly at odds with the gold standard labeling \cite{brown92class,clark03combining,blunsom11acl}.
While the empirically observed expectation for the number of tags per word type is close to one, there are many exceptions, e.g. words that occur as both nouns and verbs ({\em opening}, {\em increase}, {\em related} etc.).

In this paper we extend the Pitman-Yor HMM tagger \cite{blunsom11acl} to explicitly include a model of the lexicon that encodes from which tags a word type may be generated.
For each word type we draw an ambiguity class which is the set of tags that it may occur with, capturing the fact that words are often ambiguous between certain tags (e.g. {\em Noun} and {\em Verb}), while rarely between others (e.g. {\em Determiner} and {\em Verb}).
We extend the type based Sequential Monte Carlo (SMC) inference algorithm of \newcite{Dubbin2012Unsupervised} to incorporate our model of the lexicon, removing the need for the heuristic inference technique of \newcite{blunsom11acl}.


We start in Section \ref{sec:pyphmm} by introducing the original PYP-HMM model and our extended model of the lexicon.
Section \ref{sec:ParticleFilter} introduces a Particle Gibbs sampler for this model, a basic SMC method that generates samples from the model's posterior. 
We evaluate these algorithms in Section \ref{sec:results}, analyzing their behavior in comparisons to previously proposed state-of-the-art approaches.

\section{Background}
\label{sec:Background}

From the early work in the 1990's, much of the focus on unsupervised PoS induction has been on hidden Markov Models (HMM) \cite{brown92class,kupiec1992robust,merialdo1993taggingenglish}.  The HMM has proven to be a powerful model of PoS tag assignment.  Successful approaches generally build upon the HMM model by expanding its context and smoothing the sparse data.  Constraints such as tag dictionaries simplify inference by restricting the number of tags to explore for each word \cite{goldwater07pos}.  Ganchev et al.~\shortcite{ganchev10posterior} used posterior regularization to ensure that word types have a sparse posterior distribution over tags.  A similar approach constrains inference to only explore tag assignments such that all tokens of the same word type are assigned the same tag.  These constraints reduce tag assignment ambiguity while also providing a bias towards the natural sparsity of tag distributions in language \cite{clark03combining}. However they do not provide a model based solution to tag ambiguity.

Recent work encodes similar sparsity information with non-parametric priors, relying on Bayesian inference to achieve strong results without any tag dictionaries or constraints \cite{goldwater07pos,johnson07pos,gao08comparison}.  Liang et al.~\shortcite{liang10type} propose a type-based approach to this Bayesian inference similar to Brown et al.~\shortcite{brown92class}, suggesting that there are strong dependencies between tokens of the same word-type. Lee et al.~\shortcite{lee10dodgy} demonstrate strong results with a similar model and the introduction of a one-tag-per-type constraint on inference.

Blunsom and Cohn~\shortcite{blunsom11acl} extend the Bayesian inference approach with a hierarchical non-parametric prior that expands the HMM context to trigrams.  However, the hierarchical non-parametric model adds too many long-range dependencies for the type-based inference proposed earlier.  The model produces state-of-the art results with a one-tag-per-type constraint, but even with this constraint the tag assignments must be roughly inferred from an approximation of the expectations. 

Ambiguity classes representing the set of tags each word-type can take aid inference by making the sparsity between tags and words explicit. \newcite{toutanova08pos} showed that modelling ambiguity classes can lead to positive results with a small tag-dictionary extracted from the data.  By including ambiguity classes in the model, this approach is able to infer ambiguity classes of unknown words.

Many improvements in part-of-speech induction over the last few years have come from the use of semi-supervised approaches in the form of projecting PoS constraints across languages with parallel corpora \cite{Das2011Unsupervised} or extracting them from the wiktionary \cite{Li2012Wiki}.  These semi-supervised methods ultimately rely on a strong unsupervised model of PoS as their base.  Thus, further improvements in unsupervised models, especially in modelling tag constrains, should lead to improvements in semi-supervised part-of-speech induction. 

We find that modelling the lexicon in part-of-speech inference can lead to more efficient algorithms that match the state-of-the-art unsupervised performance.  We also note that the lexicon model relies heavily on morphological information, and suffers without it on languages with flexible word ordering.  These results promise further improvements with more advanced lexicon models.  

\section{The Pitman-Yor Lexicon Hidden Markov Model}
\label{sec:pyphmm}

This article proposes enhancing the standard Hidden Markov Model (HMM) by explicitly incorporating a model of the lexicon that consists of word types and their associated tag ambiguity classes.  
The ambiguity class of a word type is the set of possible lexical categories to which tokens of that type can be assigned.  
In this work we aim to learn the ambiguity classes unsupervised rather than have them specified in a tag dictionary.

The Lexicon HMM (Lex-HMM) extends the Pitman-Yor HMM (PYP-HMM) described by Blunsom and Cohn~\shortcite{blunsom11acl}.  When the ambiguity class of all of the word types in the lexicon is the complete tagset, the two models are the same.

\subsection{PYP-HMM}
\label{sec:base_model}

The base of the model applies a hierarchical Pitman-Yor process (PYP) prior to a trigram hidden Markov model to jointly model the distribution of a sequence of latent word tags, $\mathbf{t}$, and word tokens, $\mathbf{w}$.  
The joint probability defined by the transition, $P_\theta(t_l|t_{n-1}, t_{n-2})$, and emission, $P_\theta(w_n|t_n)$, distributions of a trigram HMM is
\begin{equation*}
\label{eq:hmmProb}
P_\theta(\mathbf{t},\mathbf{w}) = \prod_{n=1}^{N+1}P_\theta(t_l|t_{n-1}, t_{n-2})P_\theta(w_n|t_n)
\end{equation*}
where $N = |\mathbf{t}| = |\mathbf{w}|$ and the special tag $\$$ is added to denote the sentence boundaries.  
The model defines a generative process in which the tags are selected from a transition distribution, $t_l|t_{l-1}, t_{l-2}, T$, determined by the two previous tags in their history, and the word tokens are selected from the emission distribution, $w_l|t_l, E$, of the latest tag.  
\begin{align*}
t_n&|t_{n-1}, t_{n-2}, T& &\sim T_{t_{n-1}, t_{n-2}}\\
w_n&|t_n, E& &\sim E_{t_n}
\end{align*}
The PYP-HMM draws the above multinomial distributions from a hierarchical Pitman-Yor Process prior. 
The Pitman-Yor prior defines a smooth back off probability from more complex to less complex transition and emission distributions.  In the PYP-HMM trigram model, the transition distributions form a hierarchy with trigram transition distributions drawn from a PYP with the bigram transitions as their base distribution, and the bigram transitions similarly backing off to the unigram transitions.
The hierarchical prior can be intuitively understood to smooth the trigram transition distributions with bigram and unigram distributions in a similar manner to an ngram language model \cite{teh06pyplm}. 
This back-off structure greatly reduces sparsity in the trigram distributions and is achieved by chaining together the PYPs through their base distributions:
\begin{align*}
T_{ij}&|a^T, b^T, B_i& &\sim\PYP(a^T, b^T, B_i)\\
B_i&|a^B, b^B, U& &\sim\PYP(a^B, b^B, U)\\
U&|a^U, b^U& &\sim\PYP(a^U, b^U, \text{Uniform}).\\
%
%
E_i&|a^E, b^E, C& &\sim\PYP(a^E, b^E, C_i),
\end{align*}
where $T_{ij}$, $B_i$, and $U$ are trigram, bigram, and unigram transition distributions respectively, and $C_i$ is either a uniform distribution (PYP-HMM) or a bigram character language model distribution to model word morphology (PYP-HMM+LM).

Sampling from the posterior of the hierarchical PYP is calculated with a variant of the Chinese Restaurant Process (CRP) called the Chinese Restaurant Franchise (CRF) \cite{teh06pyplm,goldwater06interpolating}.  
In the CRP analogy, each latent variable (tag) in a sequence is represented by a customer entering a restaurant and sitting at one of an infinite number of tables.  
A customer chooses to sit at a table in a restaurant according to the probability 
\begin{equation}
\label{eq:nextseat}
P(z_{n} = k|\mathbf{z}_{1:n-1}) = 
\begin{cases}
	\frac{c_{k}^--a}{n-1+b} &1\leq k\leq K^-\\
	\frac{K^-a+b}{n-1+b} &k=K^-+1
\end{cases}
\end{equation}
where $z_{n}$ is the index of the table chosen by the $n$th customer to the restaurant, $\mathbf{z}_{1:n-1}$ is the seating arrangement of the previous $n-1$ customers to enter, $c_{k}^-$ is the count of the customers at table $k$, and $K^-$ is the total number of tables chosen by the previous $n-1$ customers.  
All customers at a table share the same dish, representing the value assigned to the latent variables.  
When customers sit at an empty table, a new dish is assigned to that table according to the base distribution of the PYP.  
To expand the CRP analogy to the CRF for hierarchical PYPs, when a customer sits at a new table, a new customer enters the restaurant of the PYP of the base distribution.


Blunsom and Cohn~\shortcite{blunsom11acl} explored two Gibbs sampling methods for inference with the PYP-HMM model.  
The first individually samples tag assignments for each token.  
The second employs a tactic shown to be effective by earlier works by constraining inference to only one tag per word type (PYP-1HMM).
However marginalizing over all possible table assignments for more than a single tag is intractable.  
Blunsom and Cohn~\shortcite{blunsom11acl} approximates the PYP-1HMM tag posteriors for a particular sample according to heuristic fractional table counts. 
This approximation is shown to be particularly inaccurate for values of $a$ close to one.


\subsection{The Lexicon HMM}
\label{sec:pyp_lex}
\begin{figure}
\centering
\includegraphics[width=\linewidth]{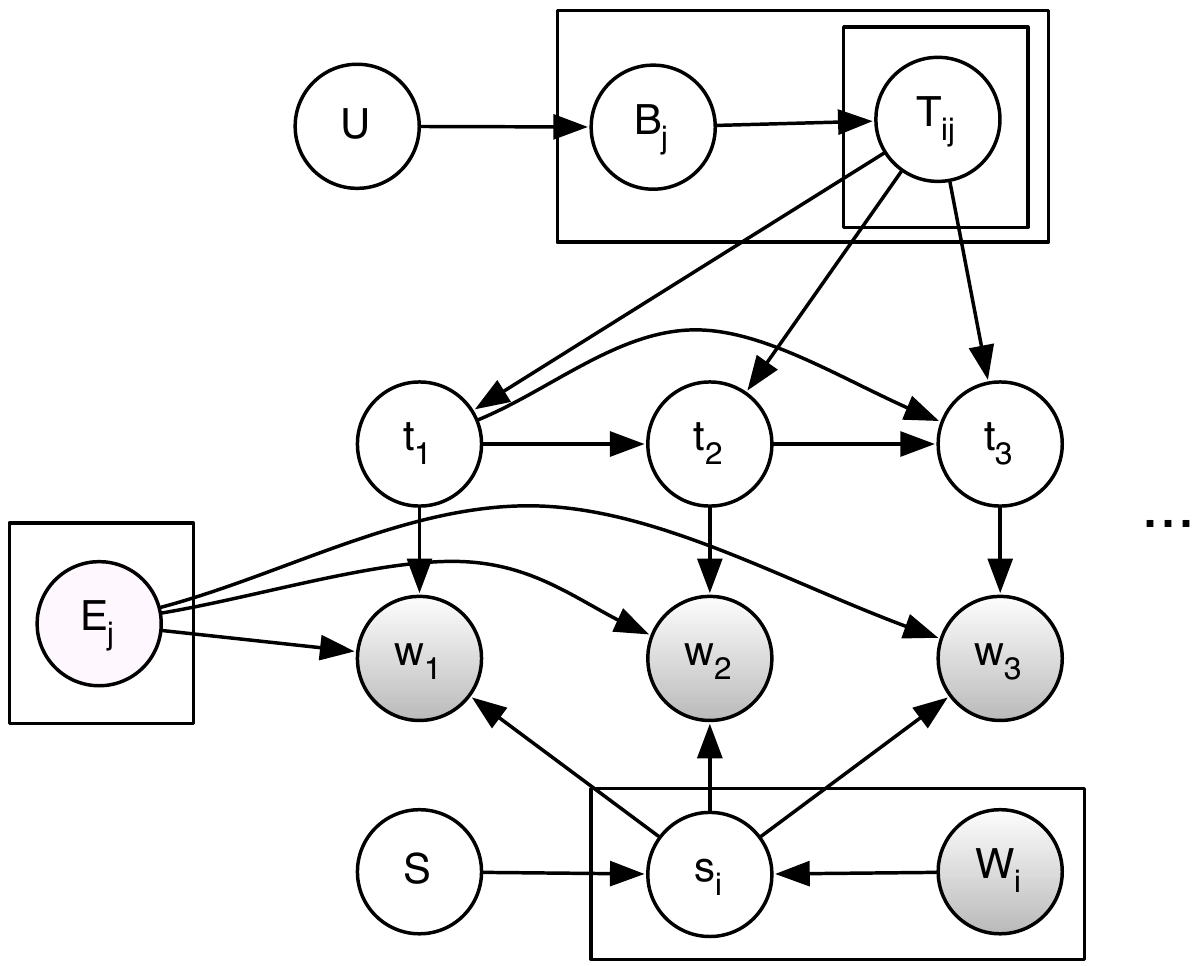}%
\caption{{\bfseries Lex-HMM Structure:} The graphical structure of the Lex-HMM model. In addition to the trigram transition ($T_{ij}$) and emission ($E_j$), the model includes an ambiguity class ($s_i$) for each word type ($W_i$) drawn from a distribution $S$ with a PYP prior.}
\label{fig:plate}
\end{figure}

We define the lexicon to be the set of all word types ($W$) and a function ($\mathcal{L}$) which maps each word type ($W_i \in W$) to an element in the power set of possible tags $T$, $$\mathcal{L} : W \rightarrow \mathcal{P}(T).$$

The Lexicon HMM (Lex-HMM) generates the lexicon with all of the word types and their ambiguity classes before generating the standard HMM parameters.  
The set of tags associated with each word type is referred to as its ambiguity class $s_i \subseteq T$.  
The ambiguity classes are generated from a multinomial distribution with a sparse, Pitman-Yor Process prior,
\begin{align*}
s_i&|S& &\sim S\\
S&|a^S, b^S& &\sim PYP(a^S,b^S, G)\\
\end{align*}
where $S$ is the multinomial distribution over all possible ambiguity classes.  The base distribution of the PYP, $G$, chooses the size of the ambiguity class according to a geometric distribution (normalized so that the size of the class is at most the number of tags $|T|$).  $G$ assigns uniform probability to all classes of the same size.
A plate diagram for this model is shown in Figure \ref{fig:plate}.

\begin{figure}
\includegraphics[width=\linewidth]{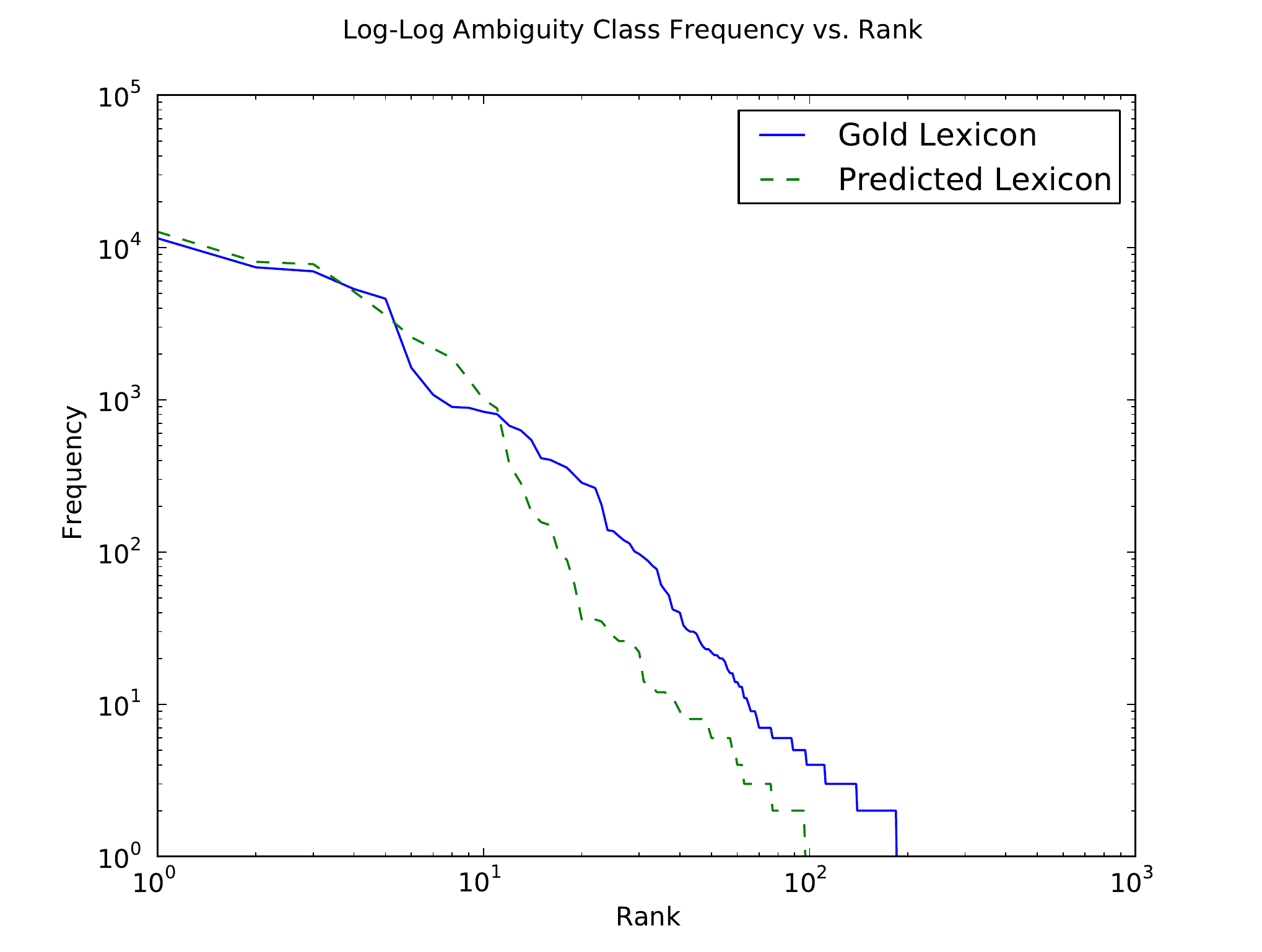}
\caption{{\bfseries Ambiguity Class Distribution:} Log-log plot of ambiguity class frequency over rank for the Penn-Treebank WSJ Gold Standard lexicon highlighting a Zipfian distribution and the ambiguity of classes extracted from the predicted tags.}
\label{fig:zipf_dists}
\end{figure}

This model represents the observation that there are relatively few distinct ambiguity classes over all of the word types in a corpus.  For example, the full Penn-Treebank Wall Street Journal (WSJ) corpus with 45 possible tags and 49,206 word types has only 343 ambiguity classes.  Figure \ref{fig:zipf_dists} shows that ambiguity classes in the WSJ have a power-law distribution.  Furthermore, these classes are generally small; the average ambiguity class in the WSJ corpus has 2.94 tags.  The PYP prior favors power-law distributions and the modified geometric base distribution favors smaller class sizes.

Once the lexicon is generated, the standard HMM parameters can be generated as described in section \ref{sec:base_model}.  The base emission probabilities $C$ are constrained to fit the generated lexicon.  The standard Lex-HMM model emission probabilities for tag $t_i$ are uniform over all word types with $t_i$ in their ambiguity class.  The character language model presents a challenge because it is non-trivial to renormalise over words with $t_i$ in their ambiguity class.  In this case word types without $t_i$ in their ambiguity class are assigned an emission probability of $0$ and the model is left deficient.

Neither of the samplers proposed by Blunsom and Cohn~\shortcite{blunsom11acl} and briefly described in section \ref{sec:base_model} are well suited to inference with the lexicon.  
Local Gibbs sampling of individual token-tag assignments would be very unlikely to explore a range of confusion classes, while the type based approximate sample relies on a one-tag-per-type restriction.
Thus in the next section we extend the Particle Filtering solution presented in \newcite{Dubbin2012Unsupervised} to the problem of simultaneous resampling the ambiguity class as well as the tags for all tokens of a given type.
This sampler provides both a more attractive inference algorithm for the original PYP-HMM and one adaptable to our Lex-HMM.

\section{Inference}
\label{sec:ParticleFilter}

To perform inference with both the lexicon and the tag assignments, we block sample the ambiguity class assignment as well as all tag assignments for tokens of the same word type.  It would be intractable to exactly calculate the probabilities to sample these blocks.  
Particle filters are an example of a Sequential Monte Carlo technique which generates unbiased samples from a distribution without summing over the intractable number of possibilities.

The particle filter samples multiple independent sequences of ambiguity classes and tag assignments.  Each sequence of samples, called a particle, is generated incrementally.  For each particle, the particle filter first samples an ambiguity class, and then samples each tag assignment in sequence based only on the previous samples in the particle.  The value of the next variable in a sequence is sampled from a proposal distribution based only on the earlier values in the sequence.  Each particle is assigned an importance weight such that a particle sampled proportional to its weight represents an unbiased sample of the true distribution.


Each particle represents a specific sampling of an ambiguity class, tag sequence, $\mathbf{t}^{W,p}_{1:n}$, and the count deltas, $\mathbf{z}^{W,p}_{1:n}$.  The term $\mathbf{t}^{W,p}_{1:n}$ denotes the sequence of $n$ tags generated for word-type $W$ and stored as part of particle $p\in[ 1,P ]$.  The count deltas store the differences in the seating arrangement neccessary to calculate the posterior probabilities according to the Chinese restaurant franchise described in section \ref{sec:base_model}.  The table counts from each particle are the only data necessary to calculate the probabilities described in equation \eqref{eq:nextseat}.

The ambiguity class for a particle is proposed by uniformly sampling one tag from the tagset to add to or remove from the previous iteration's ambiguity class with the additional possibility of using the same ambiguity class.  The particle weights are then set to 
\begin{equation*}
\frac{P(s_{W,p}|S^{-W})}{\prod_{t\in s_{W,p}}(e_t+1)^{\#(E_t)} \prod_{t \in T-s_{W,p}}(e_t)^{\#(E_t)}}
\end{equation*}
where $P(s_{W,p}|S^{-W})$ is the probability of the ambiguity class proposed for particle $p$ for word type $W$ given the ambiguity classes for the rest of the vocabulary, $e_t$ is the number of word types with $t$ in their ambiguity class, and $\#(E_t)$ is the number of tables in the CRP for the emission distribution of tag $t$.  The last two terms of the equation correct for the difference in the base probabilities of the words that have already been sampled with a different lexicon.

%
%
At each token occurrence $n$, the next tag assignment, $t^{W,p}_n$ for each particle $p\in[1,P]$ is determined by the seating decisions $z^{W,p}_n$, which are made according the proposal distribution:  
\begin{equation*}
\begin{split}
q^{W,p}_n	&(z^{W,p}_n|\mathbf{z}^{W,p}_{1:n-1},\mathbf{z}^{-W})	\propto \\
		&P(z^{W,p}_n|c^{-2},c^{-1},\mathbf{z}^{W,p}_{1:n-1},\mathbf{z}^{-W})	\\
\times&P(c^{+1}_n|c^{-1}_n,z^{W,p}_n,\mathbf{z}^{W,p}_{1:n-1},\mathbf{z}^{-W})	\\
\times&P(c^{+2}_n|z^{W,p}_n,c^{+1}_n,\mathbf{z}^{W,p}_{1:n-1},\mathbf{z}^{-W})	\\
\times&P(w^W_n|z^{W,p}_n,z^{W,p}_{1:n-1},z^{-W}).
\end{split}
\end{equation*}
In this case, $c^{\pm k}_n$ represents a tag in the context of site $t^W_n$ offset by $k$, while  $\mathbf{z}^{W,p}_{1:n-1}$ and $\mathbf{z}^{-W}$ represent the table counts from the seating decisions previously chosen by particle $p$ and the values at all of the sites where a word token of type $W$ does not appear, respectively.  This proposal distribution ignores changes to the seating arrangement between the three transitions involving the site $n$.  The specific tag assignement, $t^W,p_n$, is completely determined by the seating decisions sampled according to this proposal distribution.  Once all of the particles have been sampled, one of them is sampled with probability proportional to its weight.  This final sample is a sample from the target distribution.

As the Particle Filter is embedded in a Gibbs sampler which cycles over all word types this algorithm is an instance of Particle Gibbs.
Andrieu et al.~\shortcite{andrieu2010particlemarkov} shows that to ensure the samples generated by SMC for a Gibbs sampler have the target distribution as the invariant density, the particle filter must be modified to perform a \emph{conditional SMC update}.  This means that the particle filter guarantees that one of the final particles is assigned the same values as the previous Gibbs iteration.  Therefore, a special $0^\text{th}$ particle is automatically assigned the value from the prior iteration of the Gibbs sampler at each site $n$, though the proposal probability $q^W_n(t^{W,0}_n|\mathbf{t}^{W,p}_{1:n-1},\mathbf{z}^{W,p}_{1:n-1})$ still has to be calculated to update the weight $\omega^{W,p}_n$ properly.  This ensures that the sampler has a chance of reverting to the prior iteration's sequence.

\section{Experiments and Results}
\label{sec:results}
We provide an empirical evaluation of our proposed Lex-HMM in terms of the accuracy of the taggings learned according to the most popular metric, and the distributions over ambiguity classes.
Our experimental evaluation considers the impact of our improved Particle Gibbs inference algorithm both for the original PYP-HMM and when used for inference in our extended model. 

We intend to learn whether the lexicon model can match or exceed the performance of the other models despite focusing on only a subset of the possible tags each iteration.  We hypothesize that an accurate lexicon model and the sparsity it induces over the number of tags per word-type will improve the performance over the standard PYP-HMM model while also decreasing training time.
Furthermore, our lexicon model is novel, and its accuracy in representing ambiguity classes is an important aspect of its performance.  The model focuses inference on the most likely tag choices, represented by ambiguity classes.  

\subsection{Unsupervised Part-of-Speech Tagging}
\label{sec:LanguageResults}

\begin{table}
\centering
\begin{tabular}{lcc}
\hline
Sampler										&	M-1 Accuracy	&	Time (h)\\
\hline
Meta-Model (CGS10)&	76.1	&	--- \\
MEMM (BBDK10)			&	75.5	&	{\raise.17ex\hbox{$\scriptstyle\sim$}}40* \\
\hline
Lex-HMM									&	71.1	&	7.9 \\
Type PYP-HMM				&	70.1	&	401.2\\
Local PYP-HMM				&	70.2	&	8.6 \\
PYP-1HMM								&	75.6	&	20.6	\\
\hline												
Lex-HMM+LM						&	77.5	&	16.9	\\
Type PYP-HMM+LM		&	73.5	&	446.0	\\
PYP-1HMM+LM					&	77.5	&	34.9	\\
\hline
\end{tabular}
\caption{{\bfseries M-1 Accuracy on the WSJ Corpus:}  Comparison of the accuracy of each of the samplers with and without the language model emission prior on the English WSJ Corpus.  The second column reports run time in hours where available*.  Note the Lex-HMM+LM model matches the PYP-1HMM+LM approximation despite finishing in half the time.  The abbreviations in parentheses indicate that the results were reported in CGS10 \cite{christodoulopoulos10two} and BBDK10 \cite{bergkirkpatrick10painless} *CGS10 reports that the MEMM model takes approximately 40 hours on 16 cores.}
\label{table:WSJacc}
\end{table}

The most popular evaluation for unsupervised part-of-speech taggers is to induce a tagging for a corpus and compare the induced tags to those annotated by a linguist.
As the induced tags are simply integer labels, we must employ a mapping between these and the more meaningful syntactic categories of the gold standard.
We report results using the many-to-one (M-1) metric considered most intuitive by the evaluation of \newcite{christodoulopoulos10two}.
M-1 measures the accuracy of the model after mapping each predicted class to its most frequent corresponding tag.  While \newcite{christodoulopoulos10two} found V-measure to be more stable over the number of parts-of-speech, this effect doesn't appear when the number of tags is constant, as in our case.
For experiments on English, we report results on the entire Penn.\ Treebank \cite{marcus93penn}.  For other languages we use the corpora made available for the CoNLL-X Shared Task \cite{buchholz06conll}.  
All Lex-HMM results are reported with 10 particles as no significant improvement was found with 50 particles.


Table \ref{table:WSJacc} compares the M-1 accuracies of both the PYP-HMM and the Lex-HMM models on the Penn.\ Treebank Wall Street Journal corpus.  Blunsom and Cohn~\shortcite{blunsom11acl} found that the Local PYP-HMM+LM sampler is unable to mix, achieving accuracy below 50\%, therefore it has been left out of this analysis.  The Lex-HMM+LM model achieves the same accuracy as the state-of-the-art PYP-1HMM+LM approximation.  The Lex-HMM+LM's focus on only the most likely tags for each word type allows it to finish training in half the time as the PYP-1HMM+LM approximation without any artificial restrictions on the number of tags per type.  This contrasts with other approaches that eliminate the constraint at a much greater cost, e.g. the Type PYP-HMM, the MEMM, and the Meta-Model \footnote{While were unable to get an estimate on the runtime of the Meta-Model, it uses a system similar to the feature-based system of the MEMM with an additional feature derived from the proposed class from the brown model.  Therefore, it is likely that this model has a similar runtime.}

\begin{table*}
\small
\centering
\begin{tabular}{@{\extracolsep{\fill}}l|cccc|ccc}
\hline
{\small Language}&{\small Lex-HMM}&{\small PYP-HMM}&{\small Local}&{\small 1HMM}		&{\small Lex-HMM+LM}&{\small PYP-HMM+LM}&{\small 1HMM+LM}\\
\hline
WSJ				& 71.1 & 70.1 & 70.2 & 75.6	&	{\bfseries 77.5} &	73.5 &	{\bfseries 77.5}\\
Arabic		& 57.2 & 57.6 & 56.2 & 61.9 &	62.1 &	{\bfseries 62.7} &	62.0\\
Bulgarian	& 67.2 & 67.8 & 67.6 & 71.4 &	72.7 &	72.1 &	{\bfseries 76.2}\\
Czech			& 61.3 & 61.6 & 64.5 & 65.4 &	{\bfseries 68.2} &	67.4 &	67.9\\
Danish		& 68.6 & 70.3 & 69.1 & 70.6 &	{\bfseries 74.7} &	73.1 &	74.6\\
Dutch			& 70.3 & 71.6 & 64.1 & 73.2 & 71.7 &	71.8 &	{\bfseries 72.9} \\
Hungarian	& 57.9 & 61.8 & 64.8 & 69.6 & 64.4 &	69.9 &	{\bfseries 73.2}\\
Portuguese& 69.5 & 71.1 & 68.1 & 72.0 & 76.3 &	73.9 &	{\bfseries 77.1}\\
Spanish		& 73.2 & 69.1 & 68.5 & 74.7 & {\bfseries 80.0} &	75.2 &	78.8 \\
Swedish		& 66.3 & 63.5 & 67.6 & 67.2 & {\bfseries 70.4} &	67.6 &	68.6\\
\hline
{\small Average}		& 66.3 {\tiny $(67.2)$} & 66.5 {\tiny $(67.0)$} & 66.1 {\tiny $(66.2)$} & 70.2 {\tiny $(70.3)$} 
&	71.8 {\tiny $(72.6)$} &	70.7 {\tiny $(70.8)$} &	72.9 {\tiny $(72.9)$} \\
\hline
\end{tabular}
\caption{{\bfseries M-1 Accuracy of Lex-HMM and PYP-HMM models:} Comparison of M-1 accuracy for the lexicon based model (Lex-HMM) and the PYP-HMM model on several languages.  The Lex-HMM and PYP-HMM columns indicate the results of word type based particle filtering with 10 and 100 particles, respectively, while the Local and 1HMM columns use the token based sampler and the 1HMM approximation described by Blunsom and Cohn~\shortcite{blunsom11acl}.  The token based sampler was run for 500 iterations and the other samplers for 200.  The percentages in brakets represent the average accuracy over all languages except for Hungarian.}
\label{table:languages}
\end{table*}

The left side of table \ref{table:languages} compares the M-1 accuracies of the Lex-HMM model to the PYP-HMM model.  These models both ignore word morphology and rely on word order.  The 1HMM approximation achieves the highest average accuracy.  The Lex-HMM model matches or surpasses the type-based PYP-HMM approach in six languages while running much faster due to the particle filter considering a smaller set of parts-of-speech for each particle.  However, in the absence of morphological information, the Lex-HMM model has a similar average accuracy to the local and type-based PYP-HMM samplers.   The especially low performance on Hungarian, a language with free word ordering and strong morphology, suggests that the Lex-HMM model struggles to find ambiguity classes without morphology.  The Lex-HMM model has a higher average accuracy than the type-based or local PYP-HMM samplers when Hungarian is ignored.

The right side of table \ref{table:languages} compares the M-1 accuracies of the Lex-HMM+LM model to the PYP-HMM+LM.  The language model leads to consistently improved performance for each of the samplers excepting the token sampler, which is unable to mix properly with the additional complexity.  
The accuracies achieved by the 1HMM+LM sampler represent the previous state-of-the-art.  These results show that the Lex-HMM+LM model achieves state-of-the-art M-1 accuracies on several datasets, including the English WSJ.  The Lex-HMM+LM model performs nearly as well as, and often better than, the 1HMM+LM sampler without any restrictions on tag assignments.


The drastic improvement in the performance of the Lex-HMM model reinforces our hypothesis that morphology is critical to the inference of ambiguity classes.  Without the language model representing word morphology, the distinction between ambiguity classes is too ambiguous.  This leads the sampler to infer an excess of poor ambiguity classes.  For example, the tag assignments from the Lex-PYP model on the WSJ dataset consist of 660 distinct ambiguity classes, while the Lex-PYP+LM tag assignments only have 182 distinct ambiguity classes.

Note that while the Lex-HMM and Lex-HMM+LM samplers do not have any restrictions on inference, they do not sacrifice time.  The additional samples generated by the particle filter are mitigated by limiting the number of tags each particle must consider.  In practice, this results in the Lex-HMM samplers with 10 particles running in half time as the 1HMM samplers.  The Lex-HMM+LM sampler with 10 particles took 16.9 hours, while the 1HMM+LM sampler required 34.9 hours.  Furthermore, the run time evaluation does not take advantage of the inherent distributed nature of particle filters.  Each of the particles can be sampled completely independentally from the others, making it trivial to run each on a seperate core.

\subsection{Lexicon Analysis}
\label{sec:lex_analysis}

\begin{table*}
\small
\begin{tabular}{cc|c|p{4.4in}}
\hline
Rank	& Gold Rank	& Tags		& Top Word Types	\\
\hline
1			&	1					& NNP			& {\bfseries Mr.}, Corp.~$(1)$, Inc.~$(.99)$, Co. $(1)$, Exchange~$(.99)$ \\
2			&	2					&	NN			&	\% $(1)$, {\bfseries company}, stock $(.99)$, -RRB- $(0)$, years~$(0)$	\\
3			&	3					&	JJ			&	{\bfseries new}, {\bfseries other}, first $(.9)$, most $(0)$, major~$(1)$	\\
4			&	5					&	NNS			& {\bfseries companies}, prices $(1)$, quarter $(0)$, week $(0)$, {\bfseries investors}	\\
5			&	4					&	CD			&	\$ $(0)$, million $(1)$, {\bfseries billion}, {\bfseries 31}, \#~$(0)$	\\
\hline
\hline
15		&	303				& NN, CD	&	yen $(.47, 0)$, dollar $(1,0)$, 150 $(0,1)$, 29 $(0,1)$, 33~$(0,1)$	\\
16		&	17				&	VB, NN	&	plan $(.03,.9)$, offer $(.2, .74)$, issues $(0,0)$, increase $(.34,.66)$, end~$(.18, .81)$	\\
17		&	115				&	DT, NNP	&	As $(0, 0)$, One $(0,.01)$, First $(0,.82)$, Big $(0,.91)$, On~$(0,.01)$	\\
18		&	11				&	NN, JJ	&	market $(.99, 0)$, U.S. $(0,0)$, bank $(1,0)$, cash $(.98, 0)$, high~$(.06,.9)$	\\
20		&	22				&	VBN, JJ	&	estimated $(.58,.15)$, lost $(.43,.03)$, failed $(.35,.04)$, related $(.74,.23)$, reduced~$(.57,.12)$	\\
\hline
\end{tabular}
\caption{{\bfseries Selection of Predicted Ambiguity Classes:} Common ambiguity classes from the predicted part-of-speech assignments from the WSJ data set, and the five most common word types associated with each ambiguity class.  The sets are ranked according to the number of word types associated to them.  Words in bold are matched to exactly the same ambiguity set in the gold standard.  The lower five ambiguity classes are the most common with more than one part-of-speech.  Numbers in parentheses represent the proportion of tokens of that type assigned to each tag  in the gold standard for that ambiguity class.}
\label{table:top_csets}
\end{table*}

While section \ref{sec:LanguageResults} demonstrates that the Lex-HMM+LM sampler performs similarly to the more restricted 1HMM+LM, we also seek to evaluate the accuracy of the lexicon model itself.  We compare the ambiguity classes extracted from the gold standard and predicted tag assignments of the WSJ corpus.  We also explore the relationship between the actual and sampled ambiguity classes.

The solid curve in figure \ref{fig:zipf_dists} shows the distribution of the number of word types assigned to each ambiguity set extracted from the gold standard tag assignments from the Penn Treebank Wall Street Journal corpus.  The straight line strongly indicates that ambiguity classes follow a Zipfian distribution.  Figure \ref{fig:zipf_dists} also graphs the distribution of the ambiguity classes extracted from the best tag-assignment prediction from the model.  The predicted graph has a similar shape to the gold standard but represents half as many distinct ambiguity classes - 182 versus 343.

For a qualitative analysis of the generated lexicon, table \ref{table:top_csets} lists frequent ambiguity classes and the most common words assigned to them.  The 14 most frequent ambiguity classes contain only one tag each, the top half of table \ref{table:top_csets} shows the 5 most frequent.    One third of the word-types in the first five rows of the table are exactly matched with the ambiguity classes from the gold standard.  Most of the remaining words in those rows are assigned to a class representing almost all of the words' occurrences in the gold standard, e.g., `Corp.' is an NNP in 1514 out of 1521 occurrences.  Some words are assigned to classes with similar parts of speech, e.g. \{NNS\} rather than \{NN\} for week.

The lower half of table \ref{table:top_csets} shows the most frequent ambiguity classes with more than a single tag.  The words assigned to the \{NN,CD\}, \{DT,NNP\}, and \{NN,JJ\} classes are not themselves ambiguous.  Rather words that are unambiguously one of the two tags are often assigned to an ambiguity class with both.  The most common types in the \{NN, CD\} set are unambiguously either NN or CD.
In many cases the words are merged into broader ambiguity classes because the Lex-HMM+LM uses the language model to model the morphology of words over individual parts-of-speech, rather than entire ambiguity classes.  Therefore, a word-type is likely to be assigned a given ambiguity class as long as at least one part-of-speech in that ambiguity class is associated with morphologically similar words.  These results suggest modifying the Lex-HMM+LM to model word morphology over ambiguity classes rather than parts-of-speech.

The \{VB,NN\} and \{VBN,JJ\} are representative of true ambiguity classes.  Occurrences of words in these classes are likely to be either of the possible parts-of-speech.  These results show that the Lex-HMM is modelling ambiguity classes as intended.

\section{Conclusion}
\label{sec:conclusion}

This paper described an extension to the PYP-HMM part-of-speech model that incorporates a sparse prior on the lexicon and an SMC based inference algorithm.
These contributions provide a more plausible model of part-of-speech induction which models the true ambiguity of tag to type assignments without the loss of performance of earlier HMM models.
Our empirical evaluation indicates that this model is able to meet or exceed the performance of the previous state-of-the-art across a range of language families.

In addition to the promising empirical results, our analysis indicates that the model learns ambiguity classes that are often quite similar to those in the gold standard. 
We believe that further improvements in both the structure of the lexicon prior and the inference algorithm will lead to additional performance gains.
For example, the model could be improved by better modelling the relationship between a word's morphology and its ambiguity class.  We intend to apply our model to recent semi-supervised approaches which induce partial tag dictionaries from parallel language data \cite{Das2011Unsupervised} or the Wiktionary \cite{Li2012Wiki}.  We hypothesize that the additional data should improve the modelled lexicon and consequently improve tag assignments.

The Lex-HMM models ambiguity classes to focus the sampler on the most likely parts-of-speech for a given word-type.  In doing so, it matches or improves on the accuracy of other models while running much faster.

\bibliography{phil,greg}
\bibliographystyle{acl}

\end{document}